\documentclass{article}
\usepackage{tabularx}
\usepackage{graphicx} 
\usepackage{rotating}
\usepackage{hyperref}
\usepackage{multirow}
\usepackage{array}
\usepackage{booktabs}
\usepackage{amsmath}
\usepackage[a4paper, margin=1in]{geometry}

\title{OpenThaiGPT 1.5: A Thai-Centric Open Source \\ Large Language Model}
\author{
  Sumeth Yuenyong\thanks{sumeth.yue@mahidol.edu} \\
  \small{Department of Computer Engineering, Faculty of Engineering, Mahidol University}
  \and
  Kobkrit Viriyayudhakorn\thanks{kobkrit@iapp.co.th} \\
  \small{iApp Technology Co., Ltd.}\\
  \small{Artificial Intelligence Entrepreneur Association of Thailand (AIEAT)}
  \and
  Apivadee Piyatumrong\thanks{apivadee.pi@bdi.or.th} \\
  \small{Big Data Institute (Public Organization)}
  \and
  Jillaphat Jaroenkantasima\thanks{autsadang41@gmail.com} \\
  \small{Department of Electrical Engineering,
  Faculty of Engineering,
  Kasetsart University}
}

\date{November 2024}

\begin{document}

\maketitle

\begin{abstract}
OpenThaiGPT 1.5 is an advanced Thai language chat model based on Qwen v2.5, finetuned on over 2,000,000 Thai instruction pairs. This report provides an engineering perspective on the model's development, capabilities, and performance. We discuss the model's architecture, training process, and key features, including multi-turn conversation support, Retrieval Augmented Generation (RAG) compatibility, and tool-calling functionality. Benchmark results demonstrate OpenThaiGPT 1.5's state-of-the-art performance on various Thai language tasks, outperforming other open-source Thai language models. We also address practical considerations such as GPU memory requirements and deployment strategies.
\end{abstract}

\section{Model Architecture and Training}
\subsection{Base Model}
OpenThaiGPT 1.5 is built upon the Qwen v2.5 architecture \cite{hui2024qwen2}, leveraging its advanced capabilities as a foundation for Thai language modeling. The model is available in two sizes: 7 billion and 72 billion parameters, catering to different computational resource constraints and performance requirements. The 7B model was finetuned from Qwen/Qwen2.5-7B-Instruct on Huggingface, and the 72B model was finetuned from Qwen/Qwen2.5-72B-Instruct. Both base models have a vocabulary size of 152,064 and a maximum input length of 32,768. Inspection of the tokenizers and initial experimentation revealed that the Qwen 2.5 models already support the Thai language. For this reason, as well as due to the limitation of our computing, we opted not to perform any continued pretraining of the model with Thai data and start with instruction finetuning.

\subsection{Finetuning Process}
The model underwent extensive finetuning on a diverse dataset of over 2,000,000 Thai instruction pairs. This process was crucial in adapting the base model to the nuances of the Thai language and culture, enabling its effectiveness in handling Thai-specific domain questions. We used the NeMo\footnote{\url{https://github.com/NVIDIA/NeMo}} framework by Nvidia for finetuning. The 7B and 72B proposed models were trained using the LoRa technique \cite{hu2021lora}. The LoRa adaptor size $r$ was 64, and $\alpha$ was 128. The learning rate was $1e^{-4}$. The global batch size was 32. We used 10\% of the finetuning data as a validation set and restored the checkpoint with the lowest validation loss after training. Both models were trained on a server with 8x H100 GPUs, which was kindly provided to us by SiamAI\footnote{\url{https://siam.ai/}}. 

\subsubsection{Datasets}
The finetuning dataset for OpenThaiGPT 1.5 comprises several high-quality, diverse Thai language datasets:

\begin{itemize}
    \item \textbf{Thai Wiki Summary Dataset}: A collection of 3,000 cleaned rows generated from Wikipedia, focusing on summarization and information synthesis tasks.
    
    \item \textbf{Thai QA Multi-turn Answer Dataset}: This dataset contains 11,992 cleaned rows of multi-turn question-answering conversations, enhancing the model's ability to maintain context and generate coherent responses across multiple interactions.
    
    \item \textbf{Additional Proprietary Datasets}: Supplementary over 30 datasets covering various domains and task types, further enriching the model's knowledge and capabilities.

    \item \textbf{Synthetic Data}: We used Llama 3.1 405B\cite{dubey2024llama} to generate synthetic data by prompting it to create variations of an existing instruction, generate counterfactual questions to existing facts, and to form questions about Wikipedia articles.

    \item \textbf{LLM as a Judge}: After generating synthetic data, we used another instance of Llama 3.1 405B\cite{dubey2024llama} to review the quality of the generated data. We used in-context learning to inform the model of examples of low and high-quality instruction-answer pairs. We filtered the synthetic data to retain only instruction-answer pairs that the model judged high-quality.

    \item \textbf{English Data}: We would like the model to be English-Thai bilingual. Therefore, to alleviate catastrophic forgetting of English, we mixed in about 20\% of English data from various datasets on Huggingface such as: yahma/alpaca-cleaned, OpenAssistant/oasst1, and (a subset of) Open-Orca/OpenOrca.
\end{itemize}

These datasets were carefully curated and processed to ensure high-quality training data, contributing to the model's robust performance across a wide range of Thai language tasks and domains.

\subsection{Alignment and Safety}
We created a safety net for Reinforcement Learning from Human Feedback (RLHF) \cite{ouyang2022training} to ensure the model does not generate or respond to rude or socially sensitive topics. Our approach began with compiling a list of impolite words and sensitive subjects to which the model should refuse to answer. We manually wrote examples that included these words and addressed sensitive issues. Additionally, we employed techniques such as Easy Data Augmentation \cite{wei2019eda} to enhance our dataset, and we also experimented with jailbreaking \cite{xu2024comprehensive} some open-weight LLMs to generate more of this data locally and off-line. In total, we constructed around 5,000 records of alignment data and aligned the model using Direct Preference Optimization (DPO) \cite{rafailov2024direct}, accessible through the model aligner module of NeMo.

\section{Key Features}
The finetuning process preserved all the key features of Qwen2.5.

\subsection{Multi-turn Conversation Support}
OpenThaiGPT 1.5 preserved multi-turn conversation capability by having around 30\% of the instruction finetuning dataset consist of multi-turn conversations. This allows for more natural and coherent dialogues. This feature is essential for applications requiring extended interactions, such as customer support or interactive learning systems.

\subsection{Retrieval Augmented Generation (RAG)}
The model supports RAG, enabling it to incorporate external knowledge sources during response generation. This feature significantly enhances the model's ability to provide accurate and up-to-date information, making it suitable for applications requiring access to large, dynamic knowledge bases.

\subsection{Tool Calling Support}
OpenThaiGPT 1.5 includes a tool-calling feature that allows it to execute predefined functions or make API calls based on user queries. This capability extends the model's functionality beyond text generation, enabling it to perform tasks such as retrieving real-time data or interacting with external systems.

\subsection{Extended Context Handling}
The model can process up to 131,072 input tokens and generate up to 8,192 tokens, allowing for detailed and complex interactions. This extended context window is particularly useful for tasks requiring the analysis of long documents or generating comprehensive responses.

\section{Performance and Benchmarks}

\subsection{Evaluation Dataset}

\subsubsection{OpenThaiGPT Evaluation Dataset}
We introduce the \b{OpenThaiGPT Evaluation Dataset}, a comprehensive collection of Thai-language benchmark exams that we developed to evaluate language models' performance across various educational and professional domains. This novel dataset comprises multiple standardized tests and specialized assessments:

\begin{itemize}
    \item \textbf{Academic Standardized Tests}:
    \begin{itemize}
        \item A-Level exams (2021-2022) covering Applied Mathematics, Biology, English, and Social Studies
        \item ONET M3 (Grade 9) and M6 (Grade 12) exams from 2021-2025, including subjects like Thai, Mathematics, Social Studies, Science, and English
        \item TGAT (Thai General Aptitude Test) focusing on Critical Thinking
        \item TPAT1 (Thai Professional Aptitude Test) emphasizing Medical Ethics
    \end{itemize}
    
    \item \textbf{Professional Certifications}:
    \begin{itemize}
        \item Thai Investment Consultant Licensing Exams (IC Plain, IC Complex P2, P3)\cite{icexam}
    \end{itemize}
    
    \item \textbf{Natural Language Understanding Benchmarks}:
    \begin{itemize}
        \item Facebook Belebele Thai\cite{bandarkar-etal-2024-belebele}
        \item XCOPA Thai for cross-lingual transfer learning\cite{ponti2020xcopa}
        \item XNLI 2.0 Thai for cross-lingual natural language inference\cite{conneau2018xnli}
    \end{itemize}
\end{itemize}

The dataset features multiple-choice questions with verified answers, and includes additional annotations such as answerability flags and solution type indicators. All content has been verified by Thai speakers and is distributed under the Apache-2.0 License. The evaluation framework provides a rigorous testing environment for assessing Thai language models' capabilities across academic knowledge, professional expertise, and general language understanding.
The dataset is open-source and publicly available at Huggingface\footnote{\url{https://huggingface.co/datasets/openthaigpt/openthaigpt_eval}}, enabling researchers and developers to evaluate and compare different Thai language models using standardized benchmarks.

\subsubsection{Thai Exam Benchmark}
The Thai Exam Benchmark\cite{pipatanakul2023typhoon} is a comprehensive evaluation dataset designed to assess language models' performance on Thai educational and professional examinations. It consists of multiple-choice questions from various Thai national examinations including ONET (Ordinary National Educational Test), IC (Investment Consultant), TGAT (Thai General Aptitude Test), TPAT-1 (Thai Professional Aptitude Test), and A-Level exams. This benchmark provides a rigorous test of both language understanding and domain-specific knowledge across academic and professional contexts.

\subsubsection{M3Exam}
The M3Exam\cite{zhang2023m3exam} is a multilingual, multimodal, and multilevel benchmark for evaluating the capabilities of large language models. It includes questions from various domains, including science, mathematics, and social studies, and is designed to test the model's understanding of context, reasoning, and knowledge application.

\subsection{Evaluation Results}
OpenThaiGPT 1.5 has been evaluated on these three datasets. The results demonstrate the model's superior performance compared to other open-source Thai language models.

\begin{table}[htbp]
\centering
\small
\caption{Performance Comparison of Different Language Models in the 7B - 8B sizes on OpenThaiGPT Evaluation Dataset\\}
\begin{tabular}{l|cc|cc|cc|cc}
\hline
\multirow{2}{*}{\textbf{Exam Name}} & \multicolumn{2}{c|}{\textbf{Llama-3.1-8B}} & \multicolumn{2}{c|}{\textbf{Typhoon-v1.5x-8b}} & \multicolumn{2}{c|}{\textbf{Qwen2.5-7B}} & \multicolumn{2}{c}{\textbf{OpenThaiGPT1.5-7b}} \\
 & Correct & \% & Correct & \% & Correct & \% & Correct & \% \\
\hline
A Level & 57/120 & 47.50 & 56/120 & 46.67 & 70/120 & 58.33 & 72/120 & \textbf{60.00} \\
TGAT & 18/50 & \textbf{36.00} & 16/50 & 32.00 & 16/50 & 32.00 & 18/50 & \textbf{36.00} \\
TPAT1 & 22/40 & 55.00 & 21/40 & 52.50 & 23/40 & \textbf{57.50} & 23/40 & \textbf{57.50} \\
Investment Consult & 12/25 & 48.00 & 14/25 & 56.00 & 17/25 & 68.00 & 19/25 & \textbf{76.00} \\
Facebook Beleble TH & 146/200 & 73.00 & 156/200 & 78.00 & 158/200 & 79.00 & 162/200 & \textbf{81.00} \\
XCOPA TH & 138/200 & 69.00 & 159/200 & 79.50 & 161/200 & 80.50 & 162/200 & \textbf{81.00} \\
XNLI 2.0 TH & 110/200 & 55.00 & 113/200 & \textbf{56.50} & 106/200 & 53.00 & 109/200 & 54.50 \\
O-NET M3 Thai & 8/25 & 32.00 & 12/25 & 48.00 & 18/25 & \textbf{72.00} & 16/25 & 64.00 \\
O-NET M3 Thai & 10/20 & 50.00 & 15/20 & 75.00 & 18/20 & \textbf{90.00} & 16/20 & 80.00 \\
O-NET M3 Math & 3/16 & 18.75 & 4/16 & 25.00 & 5/16 & \textbf{31.25} & 5/16 & \textbf{31.25} \\
O-NET M3 Science & 11/26 & 42.31 & 12/26 & \textbf{46.15} & 12/26 & \textbf{46.15} & 12/26 & \textbf{46.15} \\
O-NET M3 English & 23/30 & 76.67 & 21/30 & 70.00 & 26/30 & \textbf{86.67} & 25/30 & 83.33 \\
O-NET M6 Thai & 19/65 & 29.23 & 31/65 & 47.69 & 30/65 & 46.15 & 35/65 & \textbf{53.85} \\
O-NET M6 Math & 3/17 & 17.65 & 5/17 & \textbf{29.41} & 5/17 & \textbf{29.41} & 5/17 & \textbf{29.41} \\
O-NET M6 Social & 24/55 & 43.64 & 28/55 & 50.91 & 31/55 & 56.36 & 32/55 & \textbf{58.18} \\
O-NET M6 Science & 9/28 & 32.14 & 12/28 & 42.86 & 16/28 & \textbf{57.14} & 16/28 & \textbf{57.14} \\
O-NET M6 English & 37/52 & 71.15 & 34/52 & 65.38 & 41/52 & 78.85 & 42/52 & \textbf{80.77} \\
\hline
\textbf{Total/Average} & 650/1169 & 55.60 & 709/1169 & 60.65 & 753/1169 & 64.41 & 769/1169 & \textbf{65.78} \\
\hline
\end{tabular}
\label{tab:model-comparison-7b}
\end{table}

\begin{table}[htbp]
\centering
\small
\caption{Performance Comparison of Different Language Models in the 14B sizes on OpenThaiGPT Evaluation Dataset. Note that there are no 14B sized models for Llama-3.1 and Typhoon-v1.5x.\\}
\begin{tabular}{l|ccc|ccc}
\hline
& \multicolumn{3}{c|}{\textbf{Qwen2.5-14B}} & \multicolumn{3}{c}{\textbf{OpenThaiGPT1.5-14b}}\\
\textbf{Exam Name} & Score & Total & \% & Score & Total & \% \\
\hline
A Level & 74 & 120 & 61.67 & 78 & 120 & \textbf{65.00} \\
TGAT & 22 & 50 & 44.00 & 25 & 50 & \textbf{50.00} \\
TPAT1 & 24 & 40 & \textbf{60.00} & 21 & 40 & 52.50 \\
Investment Consult & 19 & 25 & \textbf{76.00} & 18 & 25 & 72.00 \\
Facebook Beleble TH & 169 & 200 & 84.50 & 174 & 200 & \textbf{87.00} \\
XCOPA TH & 170 & 200 & 85.00 & 173 & 200 & \textbf{86.50} \\
XNLI 2.0 TH & 139 & 200 & \textbf{69.50} & 129 & 200 & 64.50 \\
O-NET M3 Thai & 19 & 25 & 76.00 & 21 & 25 & \textbf{84.00} \\
O-NET M3 Social & 18 & 20 & \textbf{90.00} & 18 & 20 & \textbf{90.00} \\
O-NET M3 Math & 7 & 16 & \textbf{43.75} & 2 & 16 & 12.50 \\
O-NET M3 Science & 13 & 26 & 50.00 & 14 & 26 & \textbf{53.85} \\
O-NET M3 English & 28 & 30 & \textbf{93.33} & 28 & 30 & \textbf{93.33} \\
O-NET M6 Thai & 34 & 65 & 52.31 & 37 & 65 & \textbf{56.92} \\
O-NET M6 Math & 4 & 17 & 23.53 & 7 & 17 & \textbf{41.18} \\
O-NET M6 Social & 33 & 55 & 60.00 & 34 & 55 & \textbf{61.82} \\
O-NET M6 Science & 14 & 28 & 50.00 & 16 & 28 & \textbf{57.14} \\
O-NET M6 English & 43 & 52 & \textbf{82.69} & 41 & 52 & 78.85 \\
\hline
\textbf{Total/Average} & 830 & 1169 & 71.09 & 836 & 1169 & \textbf{71.51} \\
\hline
\end{tabular}
\label{tab:model-comparison-14b}
\end{table}

\begin{table}[htbp]
\centering
\small
\caption{Performance Comparison of Different Language Models in the 70B - 72B sizes on OpenThaiGPT Evaluation Dataset\\}
\begin{tabular}{l|cc|cc|cc|cc}
\hline
\multirow{2}{*}{\textbf{Exam Name}} & \multicolumn{2}{c|}{\textbf{Llama-3.1-70B}} & \multicolumn{2}{c|}{\textbf{Llama-3-typhoon}} & \multicolumn{2}{c|}{\textbf{Qwen2.5-72B}} & \multicolumn{2}{c}{\textbf{OpenThaiGPT1.5-72B}} \\
 & Correct & \% & Correct & \% & Correct & \% & Correct & \% \\
\hline
A Level & 74/120 & 61.67 & 71/120 & 59.17 & 90/120 & 75.00 & 92/120 & \textbf{76.67} \\
TGAT & 20/50 & 40.00 & 23/50 & 46.00 & 24/50 & \textbf{48.00} & 23/50 & 46.00 \\
TPAT1 & 20/40 & 50.00 & 21/40 & 52.50 & 22/40 & \textbf{55.00} & 22/40 & \textbf{55.00} \\
Investment Consult & 13/25 & 52.00 & 15/25 & 60.00 & 20/25 & \textbf{80.00} & 18/25 & 72.00 \\
Facebook Beleble TH & 176/200 & 88.00 & 175/200 & 87.50 & 180/200 & \textbf{90.00} & 180/200 & \textbf{90.00} \\
XCOPA TH & 171/200 & 85.50 & 169/200 & 84.50 & 180/200 & 90.00 & 181/200 & \textbf{90.50} \\
XNLI 2.0 TH & 126/200 & 63.00 & 125/200 & 62.50 & 131/200 & 65.50 & 141/200 & \textbf{70.50} \\
O-NET M3 Thai & 14/25 & 56.00 & 19/25 & 76.00 & 19/25 & 76.00 & 21/25 & \textbf{84.00} \\
O-NET M3 Social & 19/20 & \textbf{95.00} & 19/20 & \textbf{95.00} & 18/20 & 90.00 & 19/20 & \textbf{95.00} \\
O-NET M3 Math & 4/16 & 25.00 & 7/16 & \textbf{43.75} & 6/16 & 37.50 & 6/16 & 37.50 \\
O-NET M3 Science & 16/26 & 61.54 & 14/26 & 53.85 & 17/26 & 65.38 & 19/26 & \textbf{73.08} \\
O-NET M3 English & 28/30 & 93.33 & 28/30 & 93.33 & 29/30 & \textbf{96.67} & 29/30 & \textbf{96.67} \\
O-NET M6 Thai & 39/65 & \textbf{60.00} & 36/65 & 55.38 & 39/65 & \textbf{60.00} & 37/65 & 56.92 \\
O-NET M6 Math & 10/17 & \textbf{58.82} & 7/17 & 41.18 & 4/17 & 23.53 & 7/17 & 41.18 \\
O-NET M6 Social & 42/55 & \textbf{76.36} & 37/55 & 67.27 & 35/55 & 63.64 & 36/55 & 65.45 \\
O-NET M6 Science & 16/28 & 57.14 & 14/28 & 50.00 & 18/28 & 64.29 & 19/28 & \textbf{67.86} \\
O-NET M6 English & 43/52 & 82.69 & 38/52 & 73.08 & 45/52 & 86.54 & 47/52 & \textbf{90.38} \\
\hline
\textbf{Total/Average} & 831/1169 & 71.09 & 818/1169 & 69.97 & 877/1169 & 75.02 & 897/1169 & \textbf{76.73} \\
\hline
\end{tabular}
\label{tab:model-comparison-72b}
\end{table}

For the Thai Exam Benchmark and M3Exam, OpenThaiGPT 1.5 (72B) achieved a score of 63.89\% and 70.39\% respectively, outperforming several other large language models, including some closed API-only models:

\begin{table}[h]
\centering
\begin{tabular}{l|c|c|c}
\hline
\textbf{Model} & \textbf{Average Score} & \textbf{Thai Exam Score} & \textbf{M3Exam Score} \\
\hline
Claude 3.5 Sonnet (2024-06-20) & \textbf{68.41\%} & \textbf{69.2\%} & 67.62\% \\
\textbf{OpenThaiGPT 1.5 72B} & 67.14\% & 63.89\% & \textbf{70.39\%} \\
GPT-4o (2023-05-13) & 66.26\% & 63.89\% & 68.63\% \\
Qwen2.5 72B Instruct & 63.43\% & 60.53\% & 66.33\% \\
\textbf{OpenThaiGPT 1.5 14B} & 60.41\% & 58.41\% & 62.41\% \\
Meta Llama 3.1 70B Instruct & 59.38\% & 58.23\% & 60.52\% \\
Llama 3 Typhoon v1.5x 70b Instruct & 59.34\% & 58.76\% & 59.92\% \\
\textbf{OpenThaiGPT 1.5 7B} & 53.03\% & 52.04\% & 54.01\% \\
\hline
\end{tabular}
\caption{Benchmark Results on the Thai Exam Benchmark and the M3Exam.}
\end{table}

These benchmark results highlight OpenThaiGPT 1.5's strong performance in Thai language understanding and generation tasks, positioning it as a leading option for Thai language AI applications.

\section{Conclusion}
We have developed and released OpenThaiGPT version 1.5 on Huggingface at openthaigpt/openthaigpt1.5-\{size\}b-instruct where \{size\} are 7, 14 or 72. There are based on Qwen2.5 family of models. Extensive experiments on Thai exams data showed that OpenThaiGPT1.5 is currently the most capable open model for the Thai language.

\bibliographystyle{abbrv}
\bibliography{references}

\appendix
\section{OpenThaiGPT Evaluation Dataset Examples}

Here are some examples from the OpenThaiGPT Evaluation Dataset.\\

\textbf{Example 1: Investment Consultant Exam}
\begin{quote}
Mr. A purchased units of Mutual Fund A at a price of 10.20 baht per unit. One year later, he sold those units at a price of 10.50 baht per unit. During the year, the mutual fund paid a dividend of 0.30 baht per unit. What is the rate of return that Mr. A received from his investment in Mutual Fund A? (Do not consider tax liabilities and other expenses incurred from investing in Mutual Fund A.)
\begin{itemize}
    \item (1) 2.94\%
    \item (2) 5.71\%
    \item (3) 5.88\%
    \item (4) 6.00\%
\end{itemize}
(3) is the correct answer.
\end{quote}

\textbf{Example 2: O-NET M6 Thai Language}
\begin{quote}
Which situation matches the meaning of the Thai idiom "pointing at a bird on a branch tip" (meaning to have unrealistic expectations or false hopes)?
\begin{itemize}
    \item (1) Somsak was disappointed to receive only an honorable mention in the essay contest, saying he shouldn't have pointed at a bird on a branch tip
    \item (2) Somsri studied very hard but still ranked only third in class; this is what they call pointing at a bird on a branch tip
    \item (3) Somjit has been hinting to her mother several times that she wants a mobile phone, hoping her mother will buy it after exams; Somjit is pointing at a bird on a branch tip
    \item (4) Somjai received a sum of prize money and invited friends to go shopping for pretty clothes, saying now she can finally point at a bird on a branch tip
    \item (5) Somying keeps pestering her father for a car even though she's still in junior high school; this is called pointing at a bird on a branch tip
\end{itemize}
(5) is the correct answer.
\end{quote}

\textbf{Example 3: O-NET M6 Social Studies}
\begin{quote}
When OPEC announced a reduction in oil production, causing global oil prices to rise, it affected the gasoline market, which is considered an essential commodity in Thailand. The domestic retail price of gasoline increased significantly, impacting people's lives in the country. Given this situation, if the government wants to help gasoline consumers, which measure should the government implement?
\begin{itemize}
    \item (1) Set the gasoline price at equilibrium
    \item (2) Establish lower production quotas for gasoline
    \item (3) Set a price ceiling in the gasoline market
    \item (4) Set a price floor in the gasoline market
    \item (5) Set minimum wages in the oil production industry
\end{itemize}
(3) is the correct answer.
\end{quote}

\textbf{Example 4: O-NET M6 Science}
\begin{quote}
A man with blood type AB and color blindness marries a woman with blood type O and normal vision (no family history of color blindness). They have three children together - one daughter and two sons. Which statement is correct?
\begin{itemize}
    \item (1) Both sons have normal vision
    \item (2) The children will have either blood type AB or O
    \item (3) If they have another child, it will be female
    \item (4) All three children can donate blood to their mother
    \item (5) The daughter may or may not be a carrier of color blindness
\end{itemize}
(1) is the correct answer.
\end{quote}

\textbf{Example 5: O-NET M6 Mathematics}
\begin{quote}
Given that the proposition "If Manee studies hard then Manee passes the exam" is false, and "Manee is class president or Manee passes the exam" is true, which of the following propositions must be true?
\begin{itemize}
    \item (1) Manee doesn't study hard or passes the exam
    \item (2) Manee passes the exam and is not class president
    \item (3) Manee studies hard if and only if Manee passes the exam
    \item (4) If Manee is class president then Manee doesn't study hard
    \item (5) Manee is class president if and only if Manee fails the exam
\end{itemize}
(5) is the correct answer.
\end{quote}

\textbf{Example 6: Facebook Beleble Thai\cite{bandarkar-etal-2024-belebele}}
\begin{quote}
Let your hands be as relaxed as possible while still playing all the notes correctly. Try not to move your fingers unnecessarily. This method will minimize the fatigue you experience. Remember that you don't need to press the keys harder to make the sound louder like a piano. If you want to increase the volume of the accordion, use more pressure or speed with the bellows.

Which of the following is NOT a correct tip for successfully playing the accordion?
\begin{itemize}
    \item (1) To make the sound louder, increase pressure when playing the notes
    \item (2) Try not to move fingers excessively to avoid using too much force
    \item (3) Be careful when playing notes while keeping hands as relaxed as possible
    \item (4) Increase the bellows speed to make the sound louder
\end{itemize}
(1) is the correct answer.
\end{quote}

\section{Our Links}

Official Website:
\\
\url{https://openthaigpt.aieat.or.th/} \\\newline
See our project on GitHub: 
\\
\url{https://github.com/orgs/OpenThaiGPT/repositories} \\\newline
Hugging faces:
\\
\url{https://huggingface.co/openthaigpt}\newline\\

\end{document}